\def\year{2020}\relax
\documentclass[letterpaper]{article} 
\usepackage{aaai20}  
\usepackage{times}  
\usepackage{helvet} 
\usepackage{courier}  
\usepackage[hyphens]{url}  
\usepackage{graphicx} 
\usepackage{graphicx}  
\usepackage{mathtools}
\usepackage{resizegather}
\usepackage[linesnumbered,ruled]{algorithm2e}
\usepackage{dsfont}
\usepackage{enumitem}
\usepackage{caption}
\usepackage{multirow}
\usepackage{amsmath,amssymb} 

\usepackage{amsthm}
\theoremstyle{plain}

\usepackage[section]{placeins}

\usepackage[table]{xcolor}
\definecolor{city_color_0}{rgb}{0.0,0.0,0.0}
\definecolor{city_color_1}{rgb}{0.5020,0.2510,0.5020}
\definecolor{city_color_2}{rgb}{0.9569,0.1373,0.9098}
\definecolor{city_color_3}{rgb}{0.2745,0.2745,0.2745}
\definecolor{city_color_4}{rgb}{0.4000,0.4000,0.6118}
\definecolor{city_color_5}{rgb}{0.7451,0.6000,0.6000}
\definecolor{city_color_6}{rgb}{0.6000,0.6000,0.6000}
\definecolor{city_color_7}{rgb}{0.9804,0.6667,0.1176}
\definecolor{city_color_8}{rgb}{0.8627,0.8627,0.0000}
\definecolor{city_color_9}{rgb}{0.4196,0.5569,0.1373}
\definecolor{city_color_10}{rgb}{0.5961,0.9843,0.5961}
\definecolor{city_color_11}{rgb}{0.2745,0.5098,0.7059}
\definecolor{city_color_12}{rgb}{0.8627,0.0784,0.2353}
\definecolor{city_color_13}{rgb}{1.0000,0.0000,0.0000}
\definecolor{city_color_14}{rgb}{0.0000,0.0000,0.5569}
\definecolor{city_color_15}{rgb}{0.0000,0.0000,0.2745}
\definecolor{city_color_16}{rgb}{0.0000,0.2353,0.3922}
\definecolor{city_color_17}{rgb}{0.0000,0.3137,0.3922}
\definecolor{city_color_18}{rgb}{0.0000,0.0000,0.9020}
\definecolor{city_color_19}{rgb}{0.4667,0.0431,0.1255}

\title{Importance-Aware Semantic Segmentation in Self-Driving \\with Discrete Wasserstein Training}

\author{Xiaofeng Liu{$^{1,4\dag}$},~Yuzhuo Han{$^{1,2\dag}$},~Song Bai{$^{3}$},~Yi Ge{$^{4}$},~Tianxing Wang{$^{5}$},~Xu Han{$^{6}$},\\
{\Large\textbf{Site Li{$^{4}$},~Jane You{$^{7}$},~Jun Lu{$^{1*}$}}}\\
{$^{1}$}Beth Israel Deaconess Medical Center, Harvard Medical School, Harvard University;\\{$^{2}$}School of Mathematical Sciences, Dalian University of Technology;\\{$^{3}$}Department of Statistics, University of California, Berkeley;~~~\\  {$^{4}$}Carnegie Mellon University;~~~ {$^{5}$}Fudan University;~~~ {$^{6}$}Johns Hopkins University\\{$^{7}$}Department of Computing, The Hong Kong Polytechnic University.\\
{\small{{$^{\dag}$}Contribute equally~~{$^{*}$}Corresponding Author: \tt{{jlu@bidmc.harvard.edu}}}}
}

\begin{document}

\maketitle

\begin{abstract}

Semantic segmentation (SS) is an important perception manner for self-driving cars and robotics, which classifies each pixel into a pre-determined class. The widely-used cross entropy (CE) loss-based deep networks has achieved significant progress w.r.t. the mean Intersection-over Union (mIoU). However, the cross entropy loss can not take the different importance of each class in an self-driving system into account. For example, pedestrians in the image should be much more important than the surrounding buildings when make a decisions in the driving, so their segmentation results are expected to be as accurate as possible. In this paper, we propose to incorporate the importance-aware inter-class correlation in a Wasserstein training framework by configuring its ground distance matrix. The ground distance matrix can be pre-defined following a priori in a specific task, and the previous importance-ignored methods can be the particular cases. From an optimization perspective, we also extend our ground metric to a linear, convex or concave increasing function $w.r.t.$ pre-defined ground distance. We evaluate our method on CamVid and Cityscapes datasets with different backbones (SegNet, ENet, FCN and Deeplab) in a plug and play fashion. In our extenssive experiments, Wasserstein loss demonstrates superior segmentation performance on the predefined critical classes for safe-driving.
\end{abstract}

\section{Introduction}

Semantic segmentation is an importance task in many vision-based applications or systems, such as self-driving, robotics, augmented reality and automatic surgery system \cite{yang2018image}. The goal is to densely assign class label to each pixel in the input image for precisely understanding the scene. Consequently, semantic segmentation can be treated as an image classification task at pixel level. In the past decades, significant amounts of research effort has been spent on this issue \cite{long2015fully,paszke2016enet,badrinarayanan2017segnet}.

The recent semantic segmentation method based on deep representation learning with cross-entropy (CE) loss have made considerable success on major open benchmark datasets \cite{cordts2016cityscapes,brostow2009semantic}. For each pixel in the input image, the CE loss compares the prediction with one-hot encoded ground-truth label without considering any connections to other pixels. The final loss is usually calculated as the average of the cumulative CE loss across the entire image, making each pixel contribute equally to the final loss \cite{liu2019conservative,liu2019unimodala}. This would lead to a problem for different classes with unbalanced representation in the image, due to training probably dominated by the most prevalent class.

\begin{figure}[t]
\centering
\includegraphics[width=8.5cm]{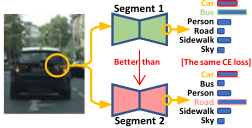}\\
\caption{The limitation of CE loss for real-world self-driving system. The ground truth class of the pixel is car ${i^*}$. Two possible softmax predictions of the segmenters have the same probability at ${i^*}$ position. Therefore, both predicted distributions have the same CE loss. However, the top prediction is preferable to the bottom, since the two predictions may result in different severity consequences.}\label{fig:1} 
\end{figure}

Even for the case that the pixels contribute unequally to the final loss, such as assigning large weights to the border of segmented objects \cite{li2017not}, the associated models still encounter challenges in practical applications.
Most existing semantic segmentation methods neglect the severity of diverse misclassifications, which may cause unexpected accidents. For example, an accident of Tesla is caused by recognising a white truck as sky, arousing intense discussion of self-vehicle safety \footnote{\url{https://www.nytimes.com/2017/01/19/business/tesla-model-s-autopilot-fatal-crash.html}}. Supposing that the white truck is recognized as a car/bus, the accident could be avoided.
Accordingly, it is necessary to investigate the severity of misclassifications in semantic segmentation method.

 Figure \ref{fig:1} shows an example to illustrate different severity consequences of misclassifications by using CE loss. For this car image, there are two possible predictions, recognising the car as bus and road by Segment1 and Segment2 respectively. The CE loss cannot discriminate these two softmax probability histograms. With one-hot ground-truth label, CE loss only depends on the prediction probability of the true class. Actually, for self-driving system,  the misclassified prediction (Car$\rightarrow$Bus) is more expected than the misclassified prediction (Car$\rightarrow$Road) in terms of severity. However, when using the CE loss, the classes are assumed to be independent of each other \cite{liu2018ordinal}. Therefore, the inter-class correlations are not properly exploited. Therefore, the inter-class correlation of (car, bus) should be closer than that of (car, road). This cannot be revealed by CE loss based models.

The importance-aware classification/segmentation \cite{chen2018importance} proposes to define some class groups based on the pre-defined importance of each class. For example, the car, truck, bus are in the most important group, road and sidewalks are in the less important group, and the sky is in the least important group. Then, a larger weight is assigned to the more important group to calculate the loss. Therefore, misclassifying a car into $any$ other classes will receive larger punishment than misclassifying the sky into $any$ other classes. Nevertheless, for a specific class, this method does not incorporate inter-class correlations between this class and any other class in the loss.

Based on the above mentioned analysis, we employ the Wasserstein loss as an alternative to empirical risk minimization. 
Specifically, we calculate the Wasserstein distance between a softmax prediction histogram and its one-hot encoded ground-truth label. By defining the ground metric based on misclassification severity, classification performance for each pixel can be measured related to inter-class correlations.

The ground metric can be predefined by regarding the severity structure as a priori, $e.g.,$ the distance between car and road is larger than car and bus. We further investigate various forms of the ground metric in optimization perspective. In the one-hot label setting, the exact Wasserstein distance can be formulated as a soft-attention scheme of all prediction probabilities and is faster computed than other general Wasserstein distance. For the semantic segmentation with unsupervised domain adaptation using constrained non-one-hot pseudo-label, we can also resort to the fast approximate solution of Wasserstein distance.

The main contributions of this paper are summarized as:

$\bullet$ We propose to render reliable segmentation results for self-driving by considering the different severity of misclassification. The inter-class correlation is explicitly incorporated as a priori to form the ground metric in our Wasserstein training framework. The importance-aware methods can be viewed as a particular case by designing a specific ground metric.

$\bullet$ For either one-hot or constrained target label in self-training-based unsupervised domain adaption setting, we systematically conclude the possible fast solution when a non-negative linear, convex or concave increasing mapping function is applied in ground metric.

$\bullet$ We empirically validate the effectiveness and generality of the proposed Wasserstein training framework which achieves promising performance on multiple challenging benchmarks with different backbone models.

\section{Related Work}

\subsection{Semantic Segmentation}

Semantic segmentation provides a comprehensive description of the scene including object category, location and shape details \cite{badrinarayanan2017segnet}. The deep learning revolution \cite{liu2018adaptive,liu2019hard,che2019deep,liu2018normalized,liu2018data} sparked wide interest in deep neural network-based semantic segmentation to replace the conventional methods \cite{liu2018joint,liu2017line}.

\cite{long2015fully} introduced a fully convolutional network for pixel or super pixel-wise classification. The conventional approaches usually employ CE loss \cite{liu2018dependency,liu2018data,liu2019feature,liu2019permutation}, which equally evaluates the errors incurred by all image pixels/classes without taking into account the different severity-level of different mistakes \cite{chen2018importance}.

The importance-aware methods \cite{chen2017importance} argue that the distinction between object/pixel importance need to be taken under consideration. The classes in Cityscapes are grouped as:

\noindent Group 4[most important]=$ \left\{{\rm Person, Car, Truck, Bus, \cdots}\right\}$;

\noindent Group 3=$ \left\{{\rm Road, Sidewalks, Train}\right\}$;

\noindent Group 2=$ \left\{{\rm Building, Wall, Fence, Vegetation, Terrain}\right\}$;

\noindent Group 1[least important]=$ \left\{{\rm Sky}\right\}$.

To compute the sum of loss in all pixels, larger weights will be given to the more important group. Consequently, the misclassification of a pixel with ground truth label in group 4 will result in a larger loss than misclassifying the sky to the other classes.

Recently, not only powerful segmentation nets \cite{chen2017deeplab} have been developed but also the pose-processing strategies are proposed to improve the initial results \cite{liu2015crf}. We note that these progress are orthogonal with our method and can be simply added to each other. 

\begin{figure*}[t]
\setlength\tabcolsep{0pt}
\begin{tabular}{cc}
\includegraphics[height=6.5cm]{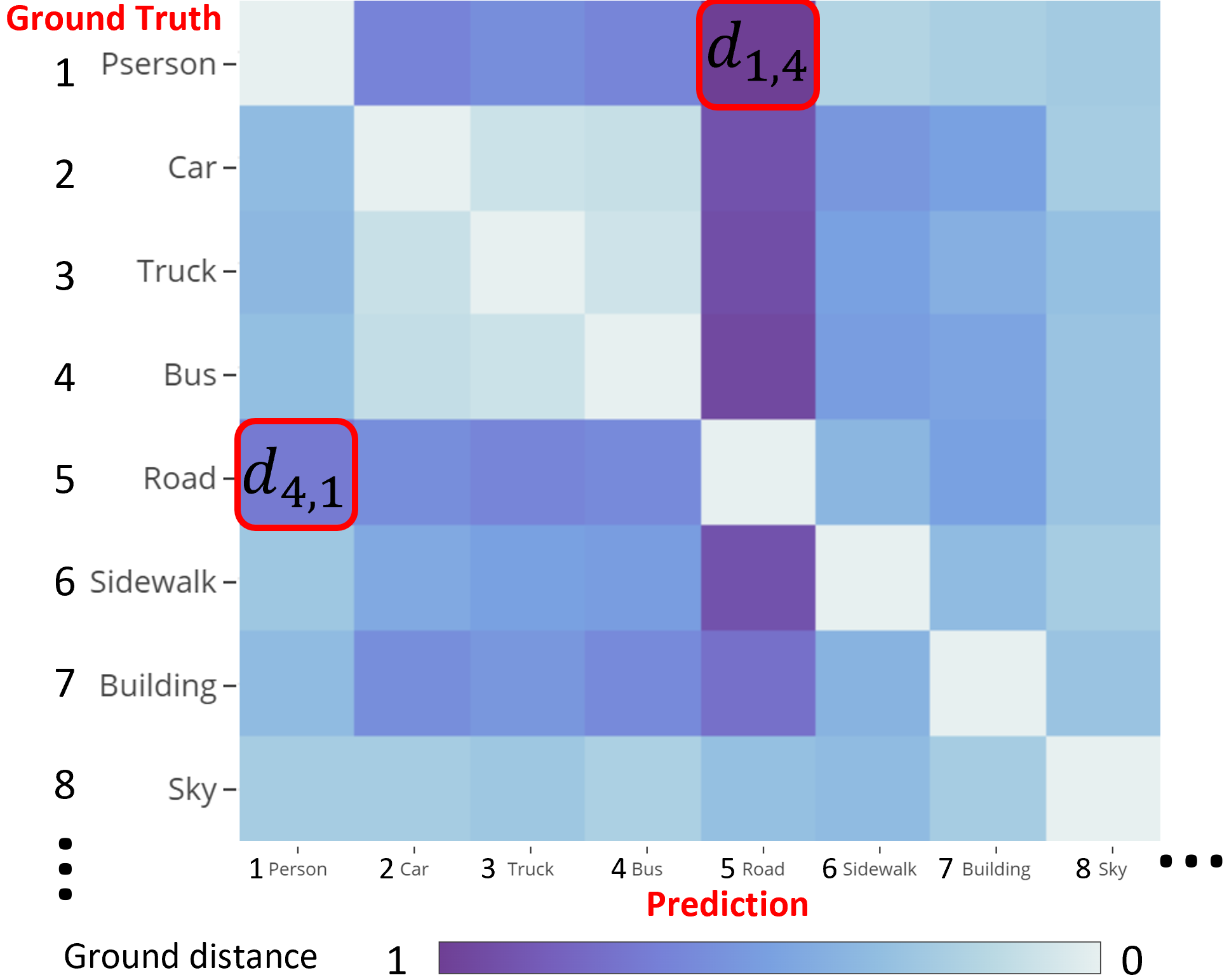}&~~~~~~~~\includegraphics[height=6.5cm]{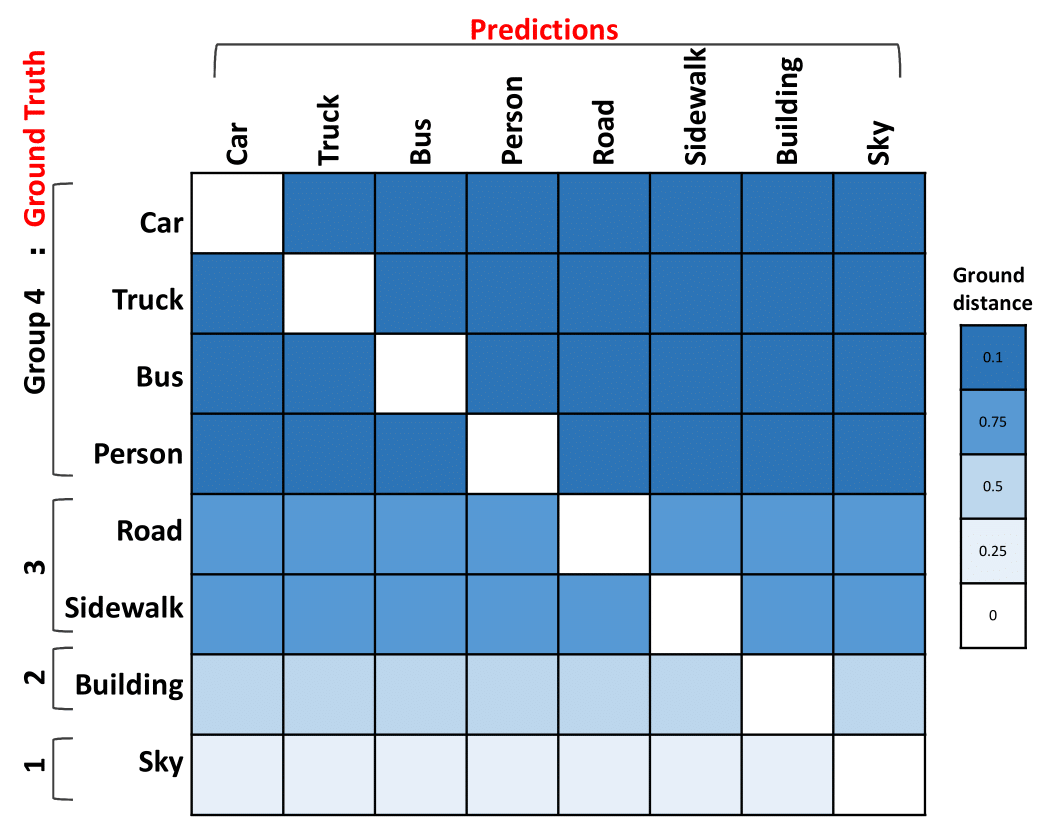}\\
\end{tabular}
\caption{Left: a possible ground matrix for severity-aware segmentation. Right: the ground matrix as an alternative for importance-aware setting.}
\label{fig:2}
\end{figure*}

\subsection{Wasserstein Distance}

{Wasserstein distance} is a measure defined between probability distributions on a given metric space \cite{kolouri2016sliced}. Recently, it has appealed to a great deal of attention in generative models $etc$ \cite{arjovsky2017wasserstein}. Due to the significant amount of computation needed to solve the exact distance for general cases, usually, it is difficult to use Wasserstein distance as a loss function. Several methods propose to solve its approximate solution, whose complexity is still in $\mathcal{O}(N^2)$ \cite{cuturi2013sinkhorn}. \cite{frogner2015learning} applies it for the multi-class multi-label task with a linear model. The fast computing of discrete Wasserstein distance is also closely related to SIFT \cite{cha2002measuring} descriptor, hue in HSV or LCH space \cite{cha2002fast} and sequence data \cite{su2017order}. 

Recently, several works propose to incorporate the Wasserstein distance as an alternative of cross-entropy loss in the context of deep learning. For example, \cite{liu2019conservative} use it for discrete and modulo classification, e.g., pose estimation. Targeting for the ordinal classification task, \cite{liu2019unimodala} propose to incorporate the correlation of health risk-level in a line to alleviate the some what sophisticate neural stick-breaking post-processing in \cite{liu2018ordinal}.

Inspired by the above works, we further adapted this idea to the severity-aware estimation, and encoded the geometry of label space by means of the ground matrix. We demonstrate that Wasserstein loss can be computed by fast algorithm in our class structure.

\section{Methodology}

The target of this work is to learn a segmenter ${h}_\theta$ which is parameterized by $\theta$. It is based on an autoencoder structure. Without loss of generality, suppose it projects a street view image ${\rm\textbf{X}}\in\mathbb{R}^{M_x\times M_x\times 3}$ to a prediction of semantic segmentation map ${\rm\textbf{S}}\in\mathbb{R}^{M_s\times M_s\times N}$, where $N$ indicates the number of categories that pre-defined by the segmentation dataset. In addition, the spatial size of input $M_x\times M_x$ and output $M_s\times M_s$ are not necessarily the same. We note that the input also not have to be the shape of square in many segmenters. Suppose ${\rm\textbf{s}}=\left\{s_i\right\}_{i=1}^{N}$ is the pixel-wise prediction of ${h}_\theta({\rm\textbf{X}})$, $i.e.,$ the $N$ classes probability normalized by softmax function. $i\in\left\{1,\cdots,{\small N}\right\}$ is the index of dimension (categories). Then we can perform learning over a hypothesis space $\mathcal{H}$ of ${h}_\theta$. Given ${\rm\textbf{X}}$ and its target one-hot ground truth label ${\rm\textbf{T}}\in\mathbb{R}^{M_s\times M_s\times N}$, typically, learning is a process by empirical risk minimization to solve $\mathop{}_{{h}_\theta\in\mathcal{H}}^{\rm min}\mathcal{L}({h}_\theta({\rm\textbf{X}}),{\rm\textbf{T}})$, with a loss $\mathcal{L}(\cdot,\cdot)$ acting as a surrogate of performance measure. In other words, it is the sum of pixel-wise error in ${M_s\times M_s}$ positions.

In the context of the self-driving risk minimization, we argue that a good loss function should reflect the properties of the importance of each class.  Unfortunately, as the previous statement, cross-entropy (CE)-based loss treat the output dimensions independently \cite{frogner2015learning}, ignoring the different severity of misclassification on label space, which is also not adaptive here. Besides, information divergence, Hellinger distance and $\mathcal{X}^2$ distance-based loss are also not the right choices, because it cannot distinguish between predictions.

Let define ${\rm\textbf{t}}=\left\{t_j\right\}_{j=1}^{N}$ as the target histogram distribution label that can be either one-hot or non-one-hot vector. Assume the class label possesses a ground metric ${\rm\textbf{D}}_{i,j}$, which measures the different severity of misclassifying $i$-th class pixel into $j$-th class pixel. There are $N^2$ possible potential outcomes ${\rm\textbf{D}}_{i,j}$ in a $N$ class dataset and form a ground distance matrix $\textbf{D}\in\mathbb{R}^{N\times N}$ \cite{ruschendorf1985wasserstein}. When ${\rm\textbf{s}}$ and ${\rm\textbf{t}}$ are both histograms, the discrete measure of exact Wasserstein loss is defined as \begin{equation}
\mathcal{L}_{\textbf{D}_{i,j}}({\rm{\textbf{s},\textbf{t}}})=\mathop{}_{\textbf{W}}^{{\rm inf}}\sum_{j=0}^{N-1}\sum_{i=0}^{N-1}\textbf{D}_{i,j}\textbf{W}_{i,j} \label{con:df}
\end{equation} where \textbf{W} is the transportation matrix with \textbf{W}$_{i,j}$ indicating the mass moved from the $i^{th}$ point in source distribution to the $j^{th}$ target position. A valid transportation matrix \textbf{W} satisfies: \\$~~~~~~~~~~~~\textbf{W}_{i,j}\geq 0$; \\$~~~~~~~~~~~~\sum_{j=0}^{N-1}\textbf{W}_{i,j}\leq s_i$;\\$~~~~~~~~~~~~\sum_{i=0}^{N-1}\textbf{W}_{i,j}\leq t_j$;\\$~~~~~~~~~~~~\sum_{j=0}^{N-1}\sum_{i=0}^{N-1}\textbf{W}_{i,j}={\rm min}(\sum_{i=0}^{N-1}s_i,\sum_{j=0}^{N-1}t_j)$.

\begin{figure*}[t]
\centering
\includegraphics[height=6cm]{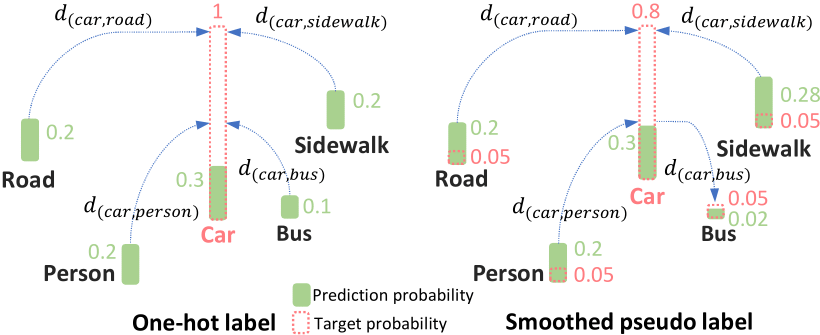}
\caption{Left: The only possible transport plan in one-hot target case. Right: the transportation in smoothed pseudo label is more complicated, $e.g.,$ car$\rightarrow$bus.}
\label{fig:3}
\end{figure*}

In mathematics, the Wasserstein or Kantorovich Rubinstein metric or distance is a distance function defined between probability distributions on a given metric space. Further, we proposes the Wasserstein distance as a loss function for unsupervised learning depends on a ground metric on the sample space of images, which is an effective distance for image retrieval, since it correlates with human perception. A possible ground distance matrix ${\rm\textbf{D}}$ which has considered different levels of importance is shown in Fig. \ref{fig:2}.

The Wasserstein distance can be the same as the Earth mover's distance when two discrete histogram distributions with the same masses ($i.e., \sum_{i=0}^{N-1}s_i=\sum_{j=0}^{N-1}t_j$) and choosing the symmetric distance $d_{i,j}$ as ${\rm\textbf{D}}_{i,j}$. However, our case is more general and different from this case. The entries in matrix ${\rm\textbf{D}}$ are not necessary to be symmetric with respect to the main diagonal. Note that the importance-aware matrix can be achieved by configuring the ground matrix as Fig. \ref{fig:2} right. The groups can be pre-defined by prior knowledge.

This setting is satisfactory for comparing the similarity of SIFT or hue, which do not use a neural network. The previous efficient algorithm usually holds only for $\textbf{D}_{i,j}={d_{i,j}}$. We propose to extend the ground metric in ${\rm\textbf{D}}_{i,j}$ as $f(d_{i,j})$, where $f$ is a positive increasing function $w.r.t.$ $d_{i,j}$.

\subsection{Wasserstein Training with One-hot Target}

In the multi-class and one-label classification tasks, the one-hot labeling is a widely-used setting. The distribution of a target label probability is ${\rm\textbf{t}}=\delta_{j,j^*}$, where $j^*$ is the ground truth class, $\delta_{j,j^*}$ is a Dirac delta, which equals to 1 for $j=j^*$\footnote{\noindent We use $i,j$ interlaced for ${\rm \textbf{s}}$ and ${\rm \textbf{t}}$, since they index the same group of positions in a label set.}, and $0$ otherwise.

\noindent\textbf{Theorem 1.} \textit{Assuming} $\sum_{j=0}^{N-1}t_j=\sum_{i=0}^{N-1}s_i$, \textit{and} ${\rm{\textbf{t}}}$ \textit{is a one-hot distribution and} $t_{j^*}=1 ($or $\sum_{i=0}^{N-1}s_i)$\footnote{We note that softmax cannot strictly guarantee the sum of its outputs to be 1 considering the rounding operation in practice. However, the difference of setting $t_{j^*}$ to $1$ or $\sum_{i=0}^{N-1}s_i)$ is not significant in our experiments using the typical format of softmax output which has up to 8 decimal places precision.}, \textit{there is only one feasible optimal transport plan.}

Following the aforementioned criteria of ${\rm\textbf{W}}$, all masses have to be transferred to the cluster of the ground truth label $j^*$, as illustrated in Fig. \ref{fig:3}. Then, the Wasserstein distance between softmax prediction {\rm{\textbf{s}}} and one-hot target {\rm{\textbf{t}}} degenerates to\begin{equation}
\mathcal{L}_{{\rm\textbf{D}}_{i,j}^{f}}({\rm{\textbf{s},\textbf{t}}})=\sum_{i=0}^{N-1} s_i f(d_{i,j^*}) 
\end{equation} We can extend the ground metric in ${\rm\textbf{D}}_{i,j}$ as $f(d_{i,j})$, where $f$ can be a linear or increasing function proper, $e.g., p^{th}$ power of $d_{i,j}$ and Huber function. The exact solution of Eq. \eqref{con:df} can be computed with a complexity of $\mathcal{O}(N)$. The ground metric term $f(d_{i,j^*})$ works as the weights $w.r.t.$ $s_i$, which takes all classes into account following a soft attention scheme \cite{liu2018dependency}. It explicitly encourages the probabilities distributing on the neighboring classes of $j^*$. Since each $s_i$ is a function of the network parameters, differentiating $\mathcal{L}_{{\rm\textbf{D}}_{i,j}^{f}} w.r.t.$ network parameters yields $\sum_{i=0}^{N-1}s_i'f(d_{i,j^*})$.

In contrast, the CE loss in one-hot setting can be formulated as $-1{\rm log}s_{j^*}$. Similar to the hard prediction scheme, only a single class prediction is considered resulting in a large information loss \cite{liu2018dependency}. Besides, the regression loss with softmax prediction could be $f(d_{i^*,j^*})$, where $i^*$ is the class with maximum prediction probability.

\begin{table*}[t]  

\renewcommand\arraystretch{1.2}
\begin{center}
\begin{tabular}{|l|c|c|c|c|c|c|c|c|}
\hline
&\multicolumn{7}{c|}{Group4}&\multirow{2}{*}{mIoU}\\\cline{2-8}

&Person&Rider&Car&Truck&Bus&Motor&Bike&\\\hline\hline

SegNet\cite{badrinarayanan2017segnet}&62.8&42.8&89.3&38.1&43.1&35.8&51.9&57.0\\\hline
+IAL\cite{chen2017importance}&84.1&46.0&91.1&75.9&65.0&22.2&\textbf{65.3}&{65.7}\\\hline
+$\mathcal{L}_{d_{i,j}}$&86.4&48.7&92.8&78.5&68.2&40.2&62.8&67.4\\\hline 
+$\mathcal{L}_{{\rm\textbf{D}}_{i,j}^2}$&87.5&\textbf{50.2}&\textbf{93.4}&\textbf{79.8}&69.5&\textbf{42.0}&64.3&\textbf{68.0}\\\hline
+$\mathcal{L}_{{\rm\textbf{D}}_{i,j}^{H\tau}}$&\textbf{87.6}&49.8&93.2&79.5&\textbf{70.3}&41.6&63.6&67.9\\\hline\hline

ENet\cite{paszke2016enet}&65.5&38.4&90.6&36.9&50.5&38.8&55.4&58.3\\\hline
+IAL\cite{chen2017importance}&87.7&41.3&92.4&\textbf{73.5}&76.2&24.1&69.7&67.5\\\hline
+$\mathcal{L}_{d_{i,j}}$&90.7&48.7&95.5&70.8&75.3&46.2&73.3&69.1\\\hline 
+$\mathcal{L}_{{\rm\textbf{D}}_{i,j}^2}$&90.9&\textbf{49.6}&\textbf{96.8}&71.4&77.6&\textbf{46.3}&\textbf{75.1}&69.3\\\hline
+$\mathcal{L}_{{\rm\textbf{D}}_{i,j}^{H\tau}}$&\textbf{90.1}&49.5&\textbf{96.8}&72.6&\textbf{77.8}&46.2&75.0&\textbf{69.5}\\\hline

\end{tabular}
\end{center}
\caption{The comparison results of various methods of Cityscapes Group 4 with SegNet and ENet backbone.}\label{tab:1}
\end{table*}

\subsection{Wasserstein Training with Conservative Target}

Deep self-training presents a powerful method for unsupervised domain adaptation in semantic segmentation, which involves an iterative process of predicting on target domain, taking the confident predictions as pseudo-labels for retraining. Obviously, self-training can put overconfident label belief on wrong classes and hence lead to deviated solutions with propagated errors because pseudo-labels can be noisy. \cite{zou2019confidence} proposes to construct the soft Pseudo-label, smoothing the one-hot Pseudo-label to a conservative target distribution. With the conservative target label, the fast computation of Wasserstein distance in Eq. \eqref{con:df} does not apply. 

Regarding it as a general case of Wasserstein distance and solving its closed-form result with a complexity higher than $\mathcal{O}(N^3)$ cannot satisfy the speed requirement of the loss function. Therefore, a possible solution is to get an approximate result with complexity in $\mathcal{O}(N^2)$. \cite{cuturi2013sinkhorn} proposes an efficient approximation of both the transport matrix and the subgradient of the loss, which is essentially a matrix balancing problem that well-studied in numerical linear algebra \cite{knight2013fast}. \cite{cuturi2013sinkhorn} uses the well-known efficient iterative Sinkhorn-Knopp algorithm.

\subsection{Monotonic Increasing $f$ w.r.t. $d_{i,j}$ as Ground Metric}

Practically, $f$ in ${\rm\textbf{D}}_{i,j}^f=f(d_{i,j})$  can be a positive increasing function $w.r.t.$ $d_{i,j}$. For simplicity the linear function is satisfactory for comparing the similarity of SIFT or hue \cite{rubner2000earth}, which even does not involve neural network optimization.

\textbf{Convex Function $ w.r.t.$ $d_{i,j}$ as Ground Metric} 

Furthermore, we can extend the ground metric as a nonnegative increasing and convex function of $d_{i,j}$. Here, we give some measures\footnote{We refer to ``measure'', since a $\rho^{th}$-root normalization is required to get a distance \cite{villani2003topics}, which satisfies three properties: positive definiteness, symmetry and triangle inequality.} using the typical convex ground metric function.

$\mathcal{L}_{{\rm\textbf{D}}_{i,j}^\rho}{(\rm{{\textbf{s},\overline{\textbf{t}}}})}$, the Wasserstein measure using $d^\rho$ as the ground metric with $\rho=2,3,\cdots$. The case $\rho=2$ is equivalent to the Cram\'{e}r distance \cite{rizzo2016energy}. Note that the Cram\'{e}r distance is not a distance metric proper. However, its square root is.

\begin{equation}
{\rm\textbf{D}}_{i,j}^\rho= d_{i,j}^\rho    
\end{equation}

$\mathcal{L}_{{\rm\textbf{D}}_{i,j}^{H\tau}}{(\rm{{\textbf{s},\overline{\textbf{t}}}})}$, the Wasserstein measure using a Huber cost function with a parameter $\tau$.

\begin{equation}
{\rm\textbf{D}}_{i,j}^{H\tau}=\left\{
             \begin{array}{ll}
             d_{i,j}^2&{\rm{if}}~d_{i,j}\leq\tau\\
             \tau(2d_{i,j}-\tau)&{\rm{otherwise}}.\\
             \end{array}
             \right.
\end{equation}

\textbf{Concave Function $ w.r.t.$ $d_{i,j}$ as Ground Metric.} 

In real world applications, it may not meaningful to choose the ground metric as the nonnegative, increasing and concave function $w.r.t.$ $d_{i,j}$. Noticing that the computing time of obtaining an closed-form solution in the conservative target label case is usually not acceptable. While the step function $f(t)=\textbf{1}_{t\neq 0}$ (one everywhere except at 0) could be a special case. It can achieve the exact solution with significantly less complexity \cite{villani2003topics}. Assuming that the $f(t)=\textbf{1}_{t\neq 0}$, the Wasserstein metric between two normalized discrete histograms on $N$ bins is simplified to the $\ell_1$ distance. 

\begin{equation}
\mathcal{L}_{\textbf{1}{d_{i,j}\neq 0}}{(\rm{{\textbf{s},\overline{\textbf{t}}}})}=\frac{1}{2}\sum_{i=0}^{N-1}{|{\rm{s}}_i-{\rm{\overline{t}}}_i|}=\frac{1}{2}||{\rm{\textbf{s}}}-{\rm{\overline{\textbf{t}}}}||_1
\end{equation}

where $||\cdot||_1$ is the discrete $\ell_1$ norm. Unfortunately, its efficient computation of closed-form solution is at the cost of losing its ability to differentiate different misclassifications.

\section{Experiments}

We show the implementation details and experimental results on two typical self-driving benchmarks ($i.e.,$ Cityscapes \cite{cordts2016cityscapes} and CamVid \cite{brostow2009semantic}). To illustrate the effectiveness of each setting choice and their combinations, we give a series of elaborate ablation studies along with the standard measures. All of the networks are pre-trained with CE loss as their vanilla version. The intersection-over-union (IoU) is defined as:
\begin{equation}
{\rm IoU}=\frac{{\rm TP}}{\rm TP+FP+FN}
\end{equation} 

where TP, FP, and FN denote the numbers of true positive, false positive, and false negative pixels, respectively. Moreover, the mean IoU is the average of IoU among all classes.

\begin{table*}[t]  
\renewcommand\arraystretch{1.2}
\begin{center}
\begin{tabular}{|l|c|c|c|c|c|c|c|}
\hline
&\multicolumn{3}{c|}{Group3}&\multicolumn{3}{c|}{Group4}&\multirow{2}{*}{mIoU}\\\cline{2-7}
&Road&Sidewalk&Sign&Car&Pedestrian&Bike&\\\hline\hline
FCN\cite{long2015fully}&98.1&89.5&25.1&84.5&64.6&38.6&69.6\\\hline
+IAL\cite{chen2017importance}&96.3&91.8&21.5&82.2&69.5&57.6&71.2\\\hline
+$\mathcal{L}_{d_{i,j}}$&98.5&93.2&28.3&87.4&71.3&60.0&72.4\\\hline 
+$\mathcal{L}_{{\rm\textbf{D}}_{i,j}^2}$&\textbf{98.7}&94.6&\textbf{29.7}&89.5&73.4&\textbf{60.7}&\textbf{72.8}\\\hline
+$\mathcal{L}_{{\rm\textbf{D}}_{i,j}^{H\tau}}$&98.5&\textbf{95.0}&29.5&\textbf{89.7}&\textbf{73.5}&60.6&\textbf{72.8}\\\hline

\end{tabular}
\end{center}
\caption{The comparison results of various methods on the Group 3/4 of CamVid dataset using FCN as backbone.}\label{tab:2}
\end{table*}

\begin{figure*}[t]
\centering
\resizebox{1\textwidth}{!}{
\begin{tabular}{@{}cccccccccc@{}}
\cellcolor{city_color_1}\textcolor{white}{~~road~~} &
\cellcolor{city_color_2}~~sidewalk~~&
\cellcolor{city_color_3}\textcolor{white}{~~building~~} &
\cellcolor{city_color_4}\textcolor{white}{~~wall~~} &
\cellcolor{city_color_5}~~fence~~ &
\cellcolor{city_color_6}~~pole~~ &
\cellcolor{city_color_7}~~traffic lgt~~ &
\cellcolor{city_color_8}~~traffic sgn~~ &
\cellcolor{city_color_9}~~vegetation~~ & 
\cellcolor{city_color_0}\textcolor{white}{~~ignored~~}\\
\cellcolor{city_color_10}~~terrain~~ &
\cellcolor{city_color_11}~~sky~~ &
\cellcolor{city_color_12}\textcolor{white}{~~person~~} &
\cellcolor{city_color_13}\textcolor{white}{~~rider~~} &
\cellcolor{city_color_14}\textcolor{white}{~~car~~} &
\cellcolor{city_color_15}\textcolor{white}{~~truck~~} &
\cellcolor{city_color_16}\textcolor{white}{~~bus~~} &
\cellcolor{city_color_17}\textcolor{white}{~~train~~} &
\cellcolor{city_color_18}\textcolor{white}{~~motorcycle~~} &
\cellcolor{city_color_19}\textcolor{white}{~~bike~~}
\end{tabular}
}\\
\includegraphics[width=15cm]{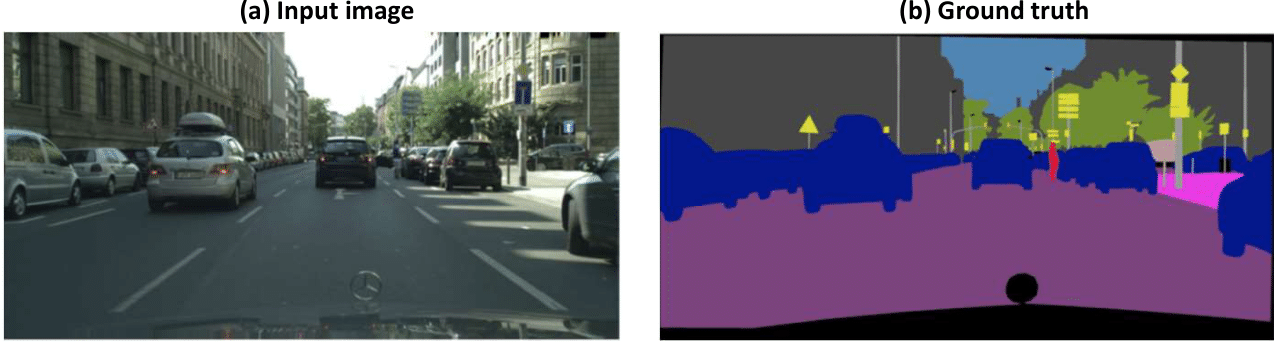}\\

\includegraphics[width=15cm]{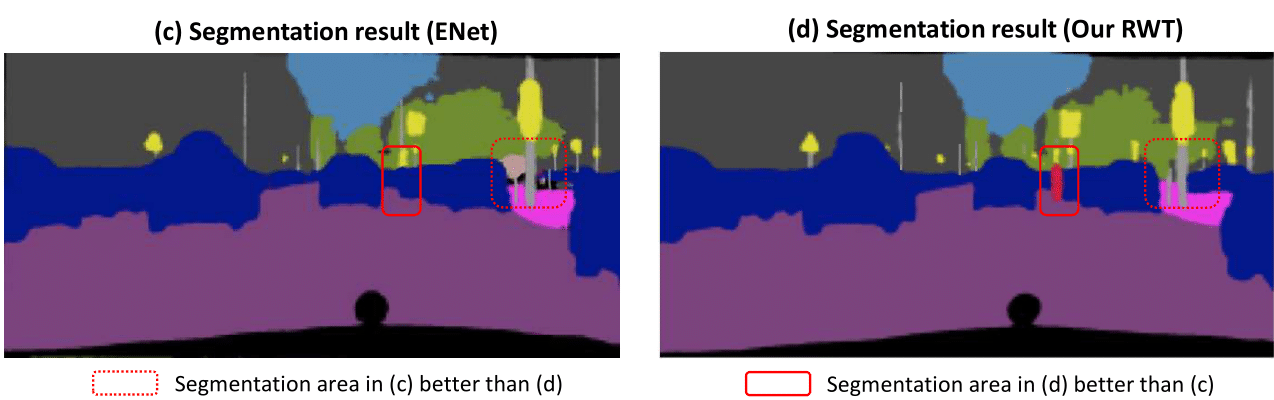}

\caption{Representative semantic segmentation result of ENet and our Wasserstein training with ENet backbone on Cityscapes dataset. The two image has the same mIoU but the misclassification of the person may lead to more severity result.}
\label{fig:6}
\end{figure*}

\subsection{Importance-aware SS with One-hot Label}
To achieve the importance-aware SS, we first pre-define our ground matrix as Fig. \ref{fig:2}. Following the setting in IAL \cite{chen2017importance,chen2018importance}, we choose the SegNet \cite{badrinarayanan2017segnet} and ENet \cite{paszke2016enet} to be our backbone. We then use IAL and our Wasserstein loss to replace the conventional CE loss in their vanilla version.

For training/validation/testing, the recent Cityscapes dataset contains 2975/500/1525 images respectively. The 19 classes that are most commonly used are selected and grouped as IAL. Table \ref{tab:1} shows that the class in group 4 are segmented with higher IoU when considering the importance of each class. Our Wasserstein loss normally outperforms 2\% than IAL, especially apply the convex function $w.r.t. d_{i,j}$. The improvements $w.r.t.$ Motor are more than 15\% over IAL.

\begin{table*}[t]  
\renewcommand\arraystretch{1.2}
\begin{center}
\begin{tabular}{|c|c|c|c|c|c|c|c|c|}
\hline
&\multicolumn{7}{c|}{Group4}&\multirow{2}{*}{mIoU}\\\cline{2-8}

&Person&Rider&Car&Truck&Bus&Motor&Bike&\\\hline\hline
LRENT\cite{zou2019confidence}&61.7&27.4&83.5&27.3&37.8&30.9&41.1&46.5\\\hline
$\mathcal{L}_{d_{i,j}}$&65.4&33.7&88.5&36.2&44.8&39.3&48.4&46.8\\\hline 
$\mathcal{L}_{{\rm\textbf{D}}_{i,j}^2}$&65.7&34.0&88.9&36.7&45.3&39.6&49.1&47.0\\\hline
$\mathcal{L}_{{\rm\textbf{D}}_{i,j}^{H\tau}}$&\textbf{66.2}&\textbf{34.7}&\textbf{89.5}&\textbf{37.1}&\textbf{46.0}&\textbf{40.8}&\textbf{50.5}&\textbf{47.3}\\\hline

\end{tabular}
\end{center}
\caption{The comparison results of various methods on the Group4 of GTA5$\rightarrow$Cityscapes unsupervised domain adaptation using DeeplabV2 as backbone.}\label{tab:3}
\end{table*}

The CamVid dataset contains 367/26/233 images for training/validation/testing respectively .We use the same setting and measurements as IAL and report the results in the table \ref{tab:2} for a fair comparison. We note that fine-tuning a public available trained FCN segmenter \cite{long2015fully} with Wasserstein loss is 1.5$\times$ faster than the training of IAL. While the IoU of some relatively unimportant classes may drop, this will have limited impact on driving safety. By introducing a stricter-than-usual objective beyond simple CE loss, we can keep the mean IoU of all classes comparable or even improved.  To intuitively present the effectiveness, we provide a representative segmentation example in Fig. \ref{fig:6}.

According to above qualitative and quantitative results, we conclude that the proposed importance-aware Wasserstein
training can improve the segmentation quality of the important objects with a large margin in terms of mIoU metric. Therefore, it is quite suitable for the application of self-driving.

\subsection{Importance-aware SS with Conservative Label}

We further test our method for unsupervised domain adaptation with constrained self-training, i.e., label entropy regularizer (LRENT) \cite{zou2019confidence}. We compute the approximate Wasserstein distance as the loss. Table \ref{tab:3} shows the performance of GTA5$\rightarrow$Cityscapes adaptation and outperforms the CE loss-based LRENT by more than 5\% in these important classes consistently. Because the Huber function is more robust to the label noise which is common for the pseudo label in self-learning method. The improvements of $\mathcal{L}_{{\rm\textbf{D}}_{i,j}^{H\tau}}$over $\mathcal{L}_{{\rm\textbf{D}}_{i,j}^2}$ are more significant than the one-hot case. This task also indicates that our approach can be considered as a general alternative objective of CE loss. Also it can be employed in a plug and play fashion.

\section{Conclusions}
Targeting for the safety driving of self-driving vehicles or robotics, we propose to implement a simple yet effective loss function for semantic segmentation based on the Wasserstein distance. It is an effective alternative of cross-entropy loss for empirical risk minimization. The importance-correlation is given by a ground metric, which can be predefined with expert knowledge. In the Wasserstein training, the importance-ignored task can be regarded simply as a special case of our importance-aware setting. Its effectiveness can be further boosted by using a convex function (e.g., square and Huber) $w.r.t. d_{i,j}$. The fast closed form solution is existed in the one-hot case, and can be used as loss function directly. Besides, the fast approximate solution can also be applied to the case with conservative label which widely exist in self-learning based unsupervised domain adaptation as well. We give extensive experiments to evidence its effectiveness and Wasserstein training achieve the state of the art performance in importance-aware tasks, and also improve the general metric mIoU with a more strictly optimization objective. Although it is originally designed for semantic segmentation tasks, we argue that our framework should has similar applicability to other problems with discrete labels that have different importance-level of misclassfication. In the future, we are planing to apply it on object detection \cite{ding2019light}, and extend it to severity-aware semantic segmentation \cite{liu2020reinforced}.

\section{Acknowledgement}
The funding support from National Institute of Health (NIH), National Institute of Neurological Disorders and Stroke (NINDS) (NS061841, NS095986), Fanhan Technology, Youth Innovation Promotion Association, CAS (2017264), Innovative Foundation of CIOMP, CAS (Y586320150) and Hong Kong Government General Research Fund GRF (Ref. No.152202/14E) are greatly appreciated.

\bibliographystyle{aaai} \bibliography{egbib}

\end{document}